# Image Compression and Watermarking scheme using Scalar Quantization

Kilari Veera Swamy [1]  B.Chandra Mohan [2]  Y.V.Bhaskar Reddy [3]  S.Srinivas Kumar [4]

[1] Professor and HOD,ECE Dept., QISCET, Ongole, India
kilarivs@yahoo.com
[2] Professor and HOD,ECE Dept., BEC, Bapatla, India
chandrabhuma@yahoo.co.in
[3] Associate Professor, ECE Dept., QISCET, Ongole, India
yvbreddy06@yahoo.com
[4] Professor, ECE Dept., JNTU, Kakinada, India
samay_ssk2@yahoo.com

*Abstract: This paper presents a new compression technique and image watermarking algorithm based on Contourlet Transform (CT). For image compression, an energy based quantization is used. Scalar quantization is explored for image watermarking. Double filter bank structure is used in CT. The Laplacian Pyramid (LP) is used to capture the point discontinuities, and then followed by a Directional Filter Bank (DFB) to link point discontinuities. The coefficients of down sampled low pass version of LP decomposed image are re-ordered in a pre-determined manner and prediction algorithm is used to reduce entropy (bits/pixel). In addition, the coefficients of CT are quantized based on the energy in the particular band. The superiority of proposed algorithm to JPEG is observed in terms of reduced blocking artifacts. The results are also compared with wavelet transform (WT). Superiority of CT to WT is observed when the image contains more contours. The watermark image is embedded in the low pass image of contourlet decomposition. The watermark can be extracted with minimum error. In terms of PSNR, the visual quality of the watermarked image is exceptional. The proposed algorithm is robust to many image attacks and suitable for copyright protection applications.*

## 1. INTRODUCTION

There are many applications requiring image compression, such as multimedia, internet, satellite imaging, remote sensing, and preservation of art work, etc. Decades of research in this area has produced a number of image compression algorithms. Most of the effort expended over the past decades on image compression has been directed towards the application and analysis of orthogonal transforms. The orthogonal transform exhibits a number of properties that make it useful. First, it generally conforms to a parseval constraint in that the energy present in the image is the same as that displayed in the image's transform.

Second, the transform coefficients bear no resemblance to the image and many of the coefficients are small and, if discarded, will provide for image compression with nominal degradation of the reconstructed image. Third, certain sub scenes of interest, such as some targets or particular textures, have easily recognized signatures in the transform domain [1]. The Discrete Cosine Transform has attracted widespread interest as a method of information coding.

The ISO/CCITT Joint Photographic Experts Group (JPEG) has selected the DCT for its baseline coding technique [2]. The fidelity loss in JPEG coding occurs entirely in quantization and much of the compression is gained by run length coding of coefficients which quantize to zero. Methods for determining the quantization table are usually based on rate-distortion theory. These methods do achieve better performance than the JPEG default quantization table [3,4,5]. However, the quantization tables are image-dependent and the complexity of the encoder is rather high. WT has attracted wide spread interest as a method of information coding [6,7]. JPEG-2000 has selected the WT for its baseline coding technique [8]. Lossless image compression algorithms using prediction are available in the literature [9,10].

Multimedia (image, video clip, text file, etc.,) data transmission over untrusted public networks is gaining importance in the recent past. For secure transaction over the internet, multimedia object has to be protected. Digital watermarking [22] is a viable solution for this problem. Digital image watermarking is a technique wherein, the watermark image is embedded in the cover image using some known watermark embedding algorithms. At the receiving side, by using known extraction algorithm, the watermark image can be obtained.

There are two fundamental requirements for any watermarking scheme. Firstly, the visual quality of the watermarked image should be good. Secondly, the watermark image should survive to many image attacks, which may be intentional or unintentional. That means, the extracted watermark should be recognizable even after modifying the watermarked image. Based on the availability of the cover image and watermark image at the receiver, the watermarking schemes are classified as blind (oblivious) and non blind (non oblivious).

Watermarking algorithms can also be classified based on the domain used for watermark embedding. Accordingly, there are two popular techniques, namely spatial domain watermarking and transform domain watermarking techniques. Spatial domain watermarking techniques are useful for higher data embedding applications. Transform domain watermarking techniques are suitable in applications where, robustness is of prime concern. There are several transform domain watermarking schemes available in the literature. Discrete Cosine Transform (DCT) [23], Discrete Fourier Transform (DFT) [24], Discrete Wavelet Transform (DWT) [25], Discrete Hadamard Transform (DHT) [26], Contourlet Transform (CT) [27], and Singular Value Decomposition(SVD) [28] are some useful transforms for image processing applications.

In this work, CT [11] is considered for image compression and watermarking. For image compression, low pass version of LP decomposed image is re-ordered and prediction algorithm is applied. Further, all coefficients of CT are quantized using scalar quantization based on the energy in that particular band. CT is applied to the entire image. Hence, blocking artifacts gets reduced in CT than JPEG compressed image. If an image contains more contours, then CT outperforms WT. The watermark image is embedded in the low pass version of LP decomposition using scalar quantization technique.

## 2. JPEG OVERVIEW

Implementation of JPEG is given below
- The image is partitioned into 8x8 blocks of pixels
- Working from left to right, top to bottm the DCT is applied to each block
- Each block is compressed using quantization
- Coding is applied for zig zag scanned coefficients

## 3. CONTOURLET TRANSFORM

CT uses double filter bank structures for obtaining sparse expansion of typical images having smooth contours [12]. In this double filter bank structure, Laplacian Pyramid (LP) is used to capture the point discontinuities, and Directional Filter Bank (DFB) is used to link these point discontinuities into linear structures. First stage is LP decomposition and second stage is DFB decomposition and is shown in Figure 1. One level LP decomposition is shown in Figure 2. Let $f(i,j)$ ($0<i, j<N$) represent the original image and its down sampled low pass filtered version by $f_{lo}(i,j)$. The prediction error is given by

$$L_o(i,j) = f(i,j) - f_{lo}'(i,j) \qquad (1)$$

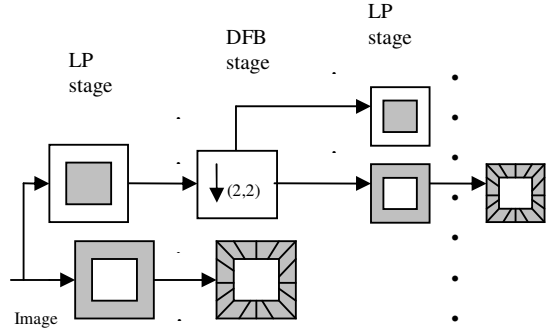

Figure 1: Contourlet Filter bank

Here, $f_{lo}'(i,j)$ represents the prediction of $f(i,j)$. Encoding step is performed on $L_o(i,j)$ as it is largely decorrelated and requires less number of bits than $f(i,j)$. In Equation 1, $L_o(i,j)$ represents a band pass image. Further encoding can be carried by applying Equation 1 on $f_{lo}(i,j)$ iteratively to get $f_{l1}(i,j)$, $f_{l2}(i,j)$,...... $f_{ln}(i,j)$, where n represents the pyramidal level number. In Figure 2, H and G represent analysis and synthesis filters respectively. M represents sampling matrix. One level LP reconstruction is shown in Figure 3. The band pass image obtained in LP decomposition is further processed by a DFB. A DFB is designed to capture the high frequency content like smooth contours and directional edges. This DFB is implemented by using a k-level binary tree decomposition that leads to $2^k$ sub bands. In CT, DFB is implemented using quincunx filter bank with fan filters [13]. The second building block of the DFB is a shearing operator, which amounts to just reordering of image samples.

CT gives two important properties [14].
- Directionality. The representation should contain much more directions.
- Anisotropy. To capture smooth contours in images, the representation contains basis elements using a variety of elongated shapes.

These two properties are useful for image compression, image watermarking, and Content Based Image Retrieval.

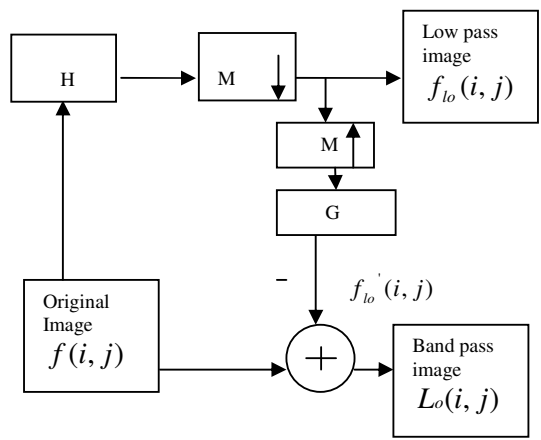

Figure 2: One level LP decomposition

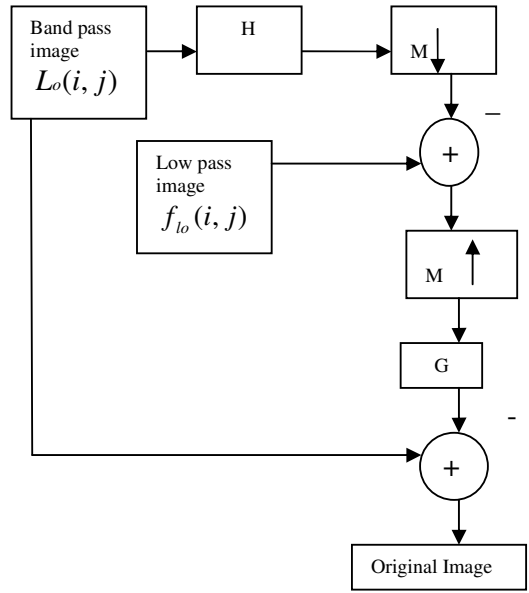

Figure 3: One level LP reconstruction

I. PROPOSED IMAGE COMPRESSION METHOD

In this method, image is decomposed using CT to obtain different bands. Down sampled low pass version of LP decomposed image is compressed in lossless procedure by steps, viz., topological re-ordering of coefficients, scanning, prediction and calculation of residues. Finally, all coefficients of CT are quantized.

### 3.1 Topological re-ordering

The rearrangement of coefficients is based purely upon coefficient position, rather than a function of coefficient value, hence it is designated as topological re-ordering [15]. The advantage of re-ordering is better accessibility of the successive coefficients for the estimation of current coefficient value. Various re-ordering possibilities are given in an earlier work [16]. Re-ordering scheme used in this work is given in Figure 4.

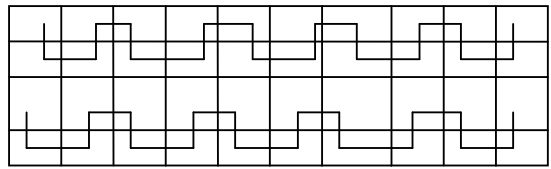

Figure 4: Square re-ordering scheme.

### 3.2 Scanning, prediction and residues

Transformed image coefficients can be processed by different scanning techniques [17]. The raster scan consists of processing the matrix by scanning the image matrix from left to right within each row and top to bottom in a matrix. Ideally, the value guessed for the current coefficient 'P' should depend on its neighbours. Notation of neighbouring coefficients of 'P' is shown in Figure 5. Prediction [18] is based on one or combination of the neighbouring coefficients. West coefficient (W) is used for estimating the current coefficient value 'P'. This is a simple technique when the coefficient values are uniform.

| NW | N | NE |
|----|---|----|
| W  | P | E  |

Figure 5: Notation of neighbouring coefficients of 'P'.

Residual is the difference between actual and estimated coefficients. Pseudo code for the algorithm is given below to find the residual matrix for low pass version of LP decomposed image.

■ *Y*= Leasterror (*X*)

Input(*X*): Low pass version of LP decomposition

Output(*Y*):Residual(current coefficient – predicted value)

*m*: rows, *n*: columns

{

Let *Y*=Ø

R=Re-order(*X*)

Scan R using scan path and at each coefficient *P* do

{

for the first coefficient in image predicted value (*Q*)=0

    *Y=P-Q*

  else predicted value (*Q*) = W

    *Y=P-W*

}

  Return *Y*

}

■

The reverse of the above procedure is considered to reconstruct the original re-ordered matrix. In both the cases, if 'P' is the first coefficient to be scanned in the beginning of the matrix then the error at 'P' is defined to be of same value. The reverse process of topological re-ordering is used to reconstruct the original matrix.

### 3.3 Quantization

Energy $\xi$ of a sub band $\varsigma(i,j)$, $0 \leq i, j \leq N$ is computed by

$$\xi = \sum_i \sum_j |\varsigma(i,j)|^2 \qquad (2)$$

Based on the energy value, each band is divided by a suitable number. The band with higher energy values is divided with less value and vice versa. All bands of CT decomposed image are quantized using scalar quantization. Quantized coefficients are rounded to get better compression. At the reconstruction, each band is multiplied with corresponding number.

## 4. PROPOSED IMAGE WATERMARKING ALGORITHM

The image watermark embedding algorithm is as follows.

- The cover image $f(i, j)$ is decomposed into low pass image and directional subbands by using CT decomposition.
- The low pass image $f_{lo}(i, j)$ coefficients are quantized using the following rule

  $z = \mod(f_{lo}(i, j), Q)$

  ➢ If $w(i, j) = 0$ & $z < Q/2$
  No modification in $f_{lo}(i, j)$

  ➢ If $w(i, j) = 0$ & $z >= Q/2$
  $f_{lo}(i, j) = f_{lo}(i, j) - Q/2$

  ➢ If $w(i, j) = 1$ & $z >= Q/2$
  No modification in $f_{lo}(i, j)$

  ➢ If $w(i, j) = 1$ & $z < Q/2$
  $f_{lo}(i, j) = f_{lo}(i, j) + Q/2$

Here, $Q$ is the quantization coefficient and may be selected based the experimentation on cover image. This is usually a trial and error process. After modifying $f_{lo}(i, j)$, inverse CT is applied with the modified $f_{lo}(i, j)$ and the watermarked image $f(i, j)'$ is obtained.

The image watermarking extraction algorithm is as follows.

- The watermarked image $f(i, j)'$ is decomposed into low pass image and directional subbands by using CT decomposition. The watermark image is extracted by using the following rule
- $z' = \mod(f_{lo}(i, j)', Q)$
  ➢ If $z < Q/2$, $w(i, j) = 0$
  ➢ If $z >= Q/2$, $w(i, j) = 1$

## 5. EXPERIMENTAL RESULTS

Experiments are performed on six grey images to verify the proposed method. These six images are represented by 8 bits/pixel and size is 512 x 512. Images used for experiments are shown in Figure 6.

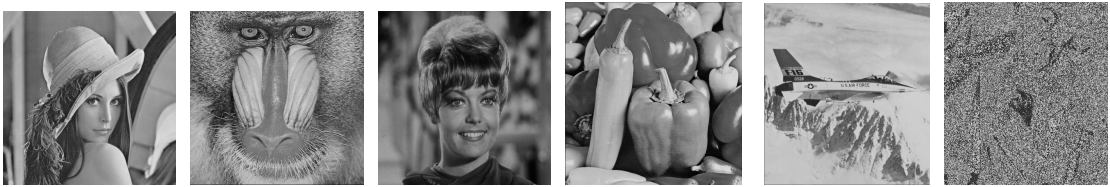

Figure 6. Images(Lena, Mandrill, Zelda, Peppers, Airplane, Barbara).

The entropy (*E*) [19] is defined as

$$E = -\sum_{e \in s} p(e) \log_2 p(e) \qquad (3)$$

where, $s$ is the set of processed coefficients and $p(e)$ is the probability of processed coefficients. By using entropy, number of bits required for compressed image is calculated. Compression ratio is defined as

$$C_R = \frac{n_1}{n_2} \qquad (4)$$

let $n_1$, $n_2$ denote the number of bits in the original and compressed images.
An often used global objective quality measure is the mean square error (MSE) defined as

$$MSE = \frac{1}{(n)(m)} \sum_{i=1}^{n} \sum_{j=1}^{m} \left[ f(i,j) - f(i,j)' \right]^2 \qquad (5)$$

where, $nxm$ is the number of total pixels. $f(i,j)$ and $f(i,j)'$ are the pixel values in the original and reconstructed image. The peak to peak signal to noise ratio (*PSNR* in dB) [20] is calculated as

$$PSNR = 10 \log_{10} \frac{255^2}{MSE} \qquad (6)$$

usable grey level values range from 0 to 255. The proposed method and JPEG methods are lossy compression techniques. JPEG process results in both blurring and blocking artifacts. Blocking effect occurs due to the discontinuity of block boundaries. The blockiness is estimated as the average differences across block boundaries. This feature is used to constitute a quality assessment model. Accordingly, the score (S) of the image [21] is calculated. The score typically has a value between 0 and 10 (10 represents the best quality, 0 the worst). The score is calculated using the formulae

$$S = \alpha + \beta B^{\gamma_1} A^{\gamma_2} Z^{\gamma_3} \qquad (7)$$

where $\alpha, \beta, \gamma_1, \gamma_2, \gamma_3$ are the model parameters that must be estimated with the subjective test data. B is related to the blockiness, A is the activity measure (relative blur) and Z is the zero-crossing (ZC) rate.

Experiments are performed with JPEG "9-7" filters because they have been shown to provide the best results for images. "9-7" filter coefficients are used both for pyramidal filter and directional filter. LP decomposition $n=(0,1,2)$ is considered for experiments. At each successive level, a number of directional sub bands are changed. Finally, LP decomposition generates 64x64 lowpass version of the original image and three 64x64, two 128x256, four 256x256 directional sub bands are generated. Energy is calculated for all bands. For Lena image, energy value of each band is 1111300, 1178, 3854, 1047, 297, 131, 20, 34, 12, 17 respectively. Hence, higher energy band is divided by less value number and lower energy value is divided by higher value number. For simplicity, lowpass band is divided with value '2', directional sub bands are divided with 4 and 8 based on the energy value. In both methods (JPEG, proposed) coding part is not considered. Entropy is calculated for rounded quantized coefficients. By using entropy number of bits required for compressed image is calculated. By using Equation 4, Compression ratio is calculated. Results are tabulated in Table 1.

Table1. Results of different images

| Image | Compression Ratio | JPEG (Score) | Proposed (Score) |
|---|---|---|---|
| Lena | 7:1 | 8.63 | 10.0 |
| Mandrill | 9:1 | 5.89 | 10.0 |
| Zelda | 10:1 | 8.04 | 10.0 |
| Peppers | 10:1 | 7.68 | 10.0 |
| Airplane | 9:1 | 8.10 | 10.0 |
| Barbara | 9:1 | 6.80 | 10.0 |

Results in Table 1, indicating that blocking artifacts are more in JPEG than the proposed method at the same compression ratio. PSNR values in dB are 37.0625 and 38.0159 for Lena and Zelda images for the proposed method. These values for JPEG are 36.9238 and 37.9238 dB. Almost at the same PSNR and compression ratio, the proposed method gives better images in terms of reduced blocking artifacts. By using WT the Score values for Mandrill and Barbara are 9.22 and 9.07 respectively. For other image-score values are same as CT. Mandrill and Barbara contains more contours. Hence, CT is more effective in capturing smooth contours and geometric structures in images than wavelet transform.

Experiments further extended to other image like Boat, Goldhill, and Nature. JPEG compression gives 5.0400, 4.8765 and 4.9357 as score values respectively. Where as the proposed method gives better score (10.0). Due to block processing, JPEG method gives more blocking artifacts than CT. CT and WT are applied for whole image. Hence, blocking artifacts are not available. Reconstructed images using three methods at 17:1 CR is shown in Figure 7. JPEG reconstructed image does not give good visual quality of image. CT and WT give same quality images. Score values for JPEG, WT and CT are 4, 10 and 10 respectively.

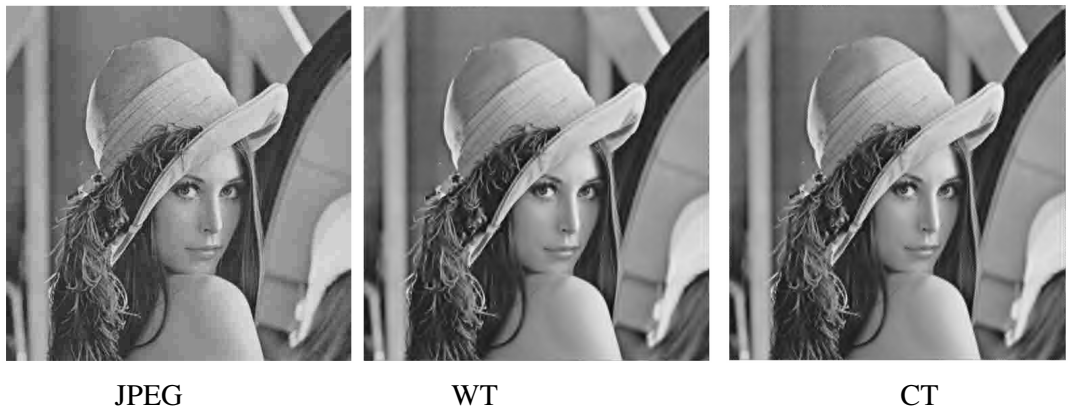

    JPEG            WT            CT

Figure 7. Reconstructed Lena images ( Compression ratio 17:1).

Experiments are conducted on LENA to test the performance of the proposed watermarking algorithm. LENA of size 512x512 is used for experimentation and is shown in Figure 8(a). The watermark image is a 32x32 binary logo as shown in Figure 8(b).

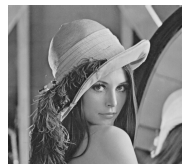

8(a)          8(b)

Figure 8(a) Cover Image Lena (b) Watermark Image

The International Journal of Next Generation Network (IJNGN), Vol.2, No.1, March 2010

Several attacks are used to assess the performance of the proposed watermarking algorithm. Two important performance metrics considered for the algorithm are PSNR (Peak Signal to Noise Ratio) and Normalized Cross correlation Coefficient (NCC) [29]. PSNR of the watermarked image is 51.12 dB indicates that the quality of the watermarked image is exceptional. Almost error free extraction is possible with the proposed algorithm. Extracted watermarks are listed in Table 2. NCC and PSNR are indicated in Table 2. Rotation is a lossy operation. The watermarked image is rotated by $10^0$ and $50^0$ to the right and then rotated back to their original position using bilinear interpolation. For low pass filtering attack, a 3x3 mask consisting of an intensity of 1/9 is used. When median filtering is applied to watermarked image, each output pixel in the attacked image contains the median value in the 3-by-3 neighbourhood around the corresponding pixel in the input image. Resizing operation first reduces or increases the size of the image and then generates the original image by using an interpolation technique. This operation is a lossy operation and hence the watermarked image also looses some watermark information. In this experiment, first the watermarked image is reduced from 512x512 size to 256x256. By using bicubic interpolation, its dimensions are increased to 512x512.

The watermarked image is compressed using lossy JPEG (Joint Photographic Experts Group) compression. The index of the JPEG compression ranges from 0 (best compression) to 100 (best quality). The watermarked image is attacked by salt & pepper noise with a noise densities of 0.001. All the extracted watermarks are clearly visible indicating the proposed method's resilience to noise attack.

Table2. Extracted Watermarks

| Rotation (5 degrees) PSNR= 20.3456 | 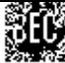 0.4515 | Resizing 34.7539 | 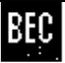 0.9624 |
|---|---|---|---|
| LPF 32.2469 | 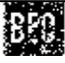 0.6035 | JPEG compression 48.0304 | 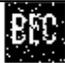 0.7945 |
| Median Filtering 36.1475 | 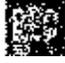 0.4329 | Salt & Pepper noise with density 0.001 | 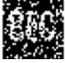 0.5298 |

6.CONCLUSIONS

In this paper, a CT based image compression and watermarking scheme is presented. LP decomposed low pass version of the image is compressed using prediction algorithm with re-ordering. Further, these coefficients are quantized using scalar quantization. All directional sub bands are quantized based on the energy of a particular sub band. Compression is improved by considering lossless compression in LL band. Due to multiresolution decomposition of CT, performance of compression is improved in terms of reduced blocking artifacts than JPEG compression. Results are also compared with wavelet transform. CT based image compression is more effective in capturing smooth contours (due to anisotropy property) than WT based compression. A CT based robust watermarking algorithm is also proposed in this work. By using scalar quantization, the watermark image is embedded in the low pass sub band of the CT decomposition. The extracted watermark is visible even after some common image attacks. Hence, it can be used for image copy right protection.

**Acknowledgment**

The first author would like to thank the management of QIS College of Engineering and Technology, Ongole, A.P, India, for their support to carry this work.

Authors:

**Dr.K.Veeraswamy** is currently working as Professor & HOD in ECE Department, QIS College of Engineering & Technology, Ongole, India. He received his M.Tech and Ph.D. from JNTU Kakinada, India. He has twelve years experience of teaching undergraduate students and post graduate students. His research interests are in the areas of image compression, image watermarking, and networking protocols.

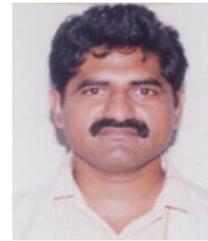

Dr.B.Chandra Mohan is currently working as Professor & HOD in ECE Department, Bapatla Engineering College, Bapatla, India. He received his Ph.D from JNTU,Kakinada, India. He received his M.Tech from Cochin University of Science & Technolgoy, Cochin, India. He has twenty years experience of teaching undergraduate students and post graduate students. His research interests are in the areas of image watermarking, and image compression.

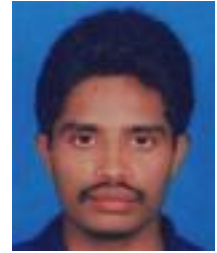

Y.V.Bhaskar reddy is presently working as an associate professor in ECE Department, QIS College of Engineering & Technology, Ongole, India. He received his B.Tech from S.V.University, Tirupathi, India and M.Tech from Bharath University, Chennai, India. He has eight years of experience of teaching undergraduate students and post graduate students. His research interests are in the areas of image processing and frequency tunable antennas.

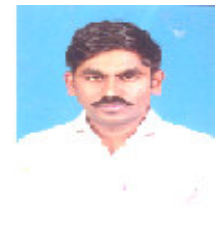

Dr. S. Srinivas Kumar is currently Professor in ECE Department and Director Sponsored Research, JNT University Kakinada, India. He received his M.Tech. from JNTU, Hyderabad, India. He received his Ph.D. from E&ECE Department IIT, Kharagpur. He has twenty three years experience of teaching undergraduate and post-graduate students and guided number of post-graduate and Ph.D thesis. He has published 20 research papers in National and International journals. Presently he is guiding five Ph.D students. His research interests are in the areas of digital image processing, computer vision, and application of artificial neural networks and fuzzy logic to engineering problems.

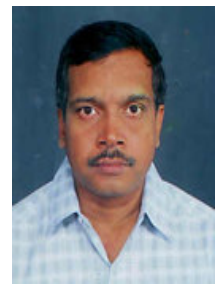